\pdfoutput=1

\documentclass[10pt, a4paper]{article}

\usepackage[final]{lrec2026} 
\usepackage{amsmath}
\usepackage{soul}
\usepackage{comment}
\usepackage{caption}
\usepackage{subcaption}
\usepackage{todonotes}
\usepackage{hyperref}
\usepackage{soul}
\usepackage{booktabs}
\usepackage{graphicx}
\usepackage[skip=2pt]{caption}   
\usepackage{adjustbox}           
\usepackage{hyperref}
\usepackage{tabularx}

\usepackage{booktabs}

\usepackage{CJKutf8}   
\usepackage{alphabeta} 
\newcommand{\cjk}[1]{\begin{CJK*}{UTF8}{gbsn}#1\end{CJK*}}

\title{How Much Noise Can BERT Handle? Insights from  Multilingual Sentence Difficulty Detection}

\name{Nouran Khallaf, Serge Sharoff} 

\address{Centre for Translation, Localisation and Interpreting Studies \\
School of Languages, Cultures and Societies\\
         University of Leeds, UK\\
         \{n.khallaf,s.sharoff\}@leeds.ac.uk}

\abstract{ Noisy training data can significantly degrade the performance of language-model-based classifiers, particularly in non-topical classification tasks. In this study we designed a methodological framework to assess the impact of denoising.  More specifically, we explored a range of denoising strategies for sentence-level difficulty detection, using training data derived from document-level difficulty annotations obtained through noisy crowdsourcing.  Beyond monolingual settings, we also address cross-lingual transfer, where a multilingual language model is trained in one language and tested in another.  We evaluate several noise reduction techniques, including Gaussian Mixture Models (GMM), Co-Teaching, Noise Transition Matrices, and Label Smoothing.  Our results indicate that while BERT-based models exhibit inherent robustness to noise, incorporating explicit noise detection can further enhance performance. For our smaller dataset, GMM-based noise filtering proves particularly effective in improving prediction quality by raising the Area-Under-the-Curve score from 0.52 to 0.92, or to 0.93 when de-noising methods are combined. However, for our larger dataset, the intrinsic regularisation of pre-trained language models provides a strong baseline, with denoising methods yielding only marginal gains (from 0.92 to 0.94, while a combination of two denoising methods made no contribution).  Nonetheless, removing noisy sentences (about 20\% of the dataset) helps in producing a cleaner corpus with fewer infelicities. As a result we have released the largest multilingual corpus for sentence difficulty prediction:  see {\footnotesize \url{https://github.com/Nouran-Khallaf/denoising-difficulty}}.
 \\  
\newline 
}

\begin{document}

\maketitleabstract

\section{Introduction}
\label{sec:org549443b}

Modern Natural Language Processing (NLP) methods, driven by Pre-Trained Language Models (PLMs) and Large Language Models (LLMs), have achieved impressive results across a wide range of tasks. Nevertheless, their performance remains uneven in \textit{non-topical classification} tasks, such as predicting text genre \cite{ronnqvist22explaininggenres,kuzman2023chatgpt}, demographic properties \cite{kang19dataset}, or text difficulty \cite{north22complexity}. Such tasks require sensitivity to subtle stylistic, structural, and syntactic cues rather than reliance on topical keywords, while LLMs struggle to capture these linguistic distinctions \cite{kuzman2023chatgpt}.  

In contrast to topical classification, which benefits from the presence of explicit domain-related terms, non-topical classification demands that models identify latent stylistic features and register-level indicators \cite{dewdney01}. Both PLMs and LLMs can be misled by irrelevant topical content in these settings \cite{roussinov23emnlp}. Moreover, recent studies demonstrate that larger generative LLMs such as GPT or LLaMA do not consistently outperform smaller PLMs such as BERT on text classification tasks \cite{edwards24incontext}. Accordingly, our experiments focus on fine-tuning BERT-like PLMs; we do not include LLM-based classifiers due to their high computational costs and marginal benefits for this task (see Section 10 for discussion of computational and ethical considerations).

\begin{table*}[t]
\centering
\small
\setlength{\tabcolsep}{6pt}
\renewcommand{\arraystretch}{1.15}
\begin{tabular}{l|r|rrc|rrc}
\hline
\textbf{Language} & \textbf{\#Docs} & \multicolumn{3}{c|}{\textbf{Wikipedia}}  & \multicolumn{3}{c}{\textbf{Vikidia}} \\ \cline{3-8}
 &  & \textbf{\#Words} & \textbf{\#Sentences} & \textbf{IQR} & \textbf{\#Words} & \textbf{\#Sentences} & \textbf{IQR} \\ \hline
Catalan & 179   & 396{,}277   & 16{,}813      & (15, 29)  & 19{,}394   & 1{,}000      & (13, 23)  \\
English & 2585   & 8{,}281{,}625   & 340{,}924      & (16, 29)  & 424{,}306   & 22{,}462      & (13, 22)  \\
Spanish & 3875   & 7{,}946{,}169   & 301{,}241      & (16, 33)  & 607{,}990   & 27{,}825      & (14, 26)  \\
French & 33438   & 46{,}618{,}143   & 1{,}945{,}046      & (15, 29)  & 6{,}643{,}567   & 320{,}372      & (14, 25)  \\
Italian & 3902   & 6{,}790{,}163   & 263{,}271      & (16, 32)  & 537{,}723   & 25{,}202      & (14, 26)  \\
Russian & 136   & 334{,}416   & 17{,}618      & (13, 22)  & 10{,}454   & 604      & (12, 20)  \\
 \hline
\end{tabular}%

\caption{Statistics for paired Wikipedia--Vikidia corpora. \#Docs is the document count (paired Wikipedia--Vikidia articles by topic). \#Words and \#Sentences are totals across all documents in each subset (restricted to paired topics). IQR gives the inter-quartile range of sentence lengths (in words) in each subset.}
\label{tabData}
\end{table*}

A major challenge in non-topical classification tasks lies in the scarcity of high-quality, reliable training data, especially in multilingual contexts. Existing datasets frequently suffer from annotation noise and label inconsistencies.  Some of this noise stems from the variability of crowd-sourced annotations and from the use of document-level difficulty labels for sentence-level predictions. For example, a sentence taken from a Wikipedia article (the default source of more complex texts in our experiments) may in fact be linguistically simple, while some sentences from an ostensibly ``simple'' crowd-sourced corpus may be structurally or semantically challenging because of annotation inconsistencies. This can seriously degrade classifier performance and obscure genuine cross-lingual trends.  

In this study, we systematically analyse the effects of annotation noise on sentence-level difficulty classification by exploring a range of denoising techniques aimed at improving robustness under realistic data conditions. With this paper we:
\begin{enumerate}
    \item identify effective noise reduction methods that enhance the stability and accuracy of non-topical classification;  
    \item evaluate intersection of data points identified as noise by several noise reduction methods;
    \item evaluate the interaction between noise and cross-lingual transfer, thus providing evidence for how multilingual PLMs handle noise across languages; and  
    \item release a publicly available de-noised multilingual corpus for sentence-level difficulty classification, accompanied by all scripts and models for reproducible experimentation.
\end{enumerate}

\section{Dataset}
\label{sec:orgbd1a179}
The datasets used in our experiments were sourced from Vikidia and Wikipedia,\footnote{The datasets from Wikipedia and Vikidia are available under the (CC BY-SA) license. Our use aligns with their intended purpose, with our modifications limited to preprocessing and data selection.} covering multiple languages, see \autoref{tabData}.  The simple versions have been crawled from Vikidia,\footnote{\url{https://www.vikidia.org/}} a website that maintains Wikipedia-style content aimed at ``children and anyone seeking easy-to-read content''.  The range of languages in our experiments reflects the availability of languages in Vikidia and the availability of annotators for quality control.  We have removed the entries marked as stubs (with little content at the moment) and collected the corresponding main Wikipedia entries for the respective languages.  Therefore, the documents in our dataset address exactly the same topics.  This removes topic biases, which often impact non-topical classification tasks due to unreasonable performance through just learning the keywords \cite{roussinov23emnlp}.  In the end, we have obtained a document-level resource for text-difficulty detection.  However, our aim is to develop more granular classifiers on the sentence level.  This formulation introduces noise, since many sentences may be simple despite their provenance from Wikipedia. 

The Inter-Quartile Range (IQR) values in \autoref{tabData} indicate that the length of most sentences in either source falls between roughly 13 and 33 words.  Across all languages, Vikidia consistently features shorter sentences, typically \textit{2–7 words shorter}.  While the documents are paired between the Wikipedia and Vikidia in each language, the number of sentences for the classification task is severely skewed towards more complex examples, as Wikipedia documents are considerably longer, leading to having 15 times more complex-labelled sentences for English and 6 times more for French.

\section{Methodology}
\label{sec:org4236f5b}

All of our experiments aim at a binary classification task: predicting whether a sentence is complex, i.e., in our definition of complexity, the sentence is not suitable for inclusion in Vikidia.  According to Table~\ref{tabData} this is the majority class, so a classifier may naturally produce a high proportion of false positives (simple sentences predicted as complex) simply because most sentences are complex and the prior pushes predictions toward that class: a degenerate classifier which only predicts the majority class will achieve about 0.93 F1-Score.  A false negative means a complex sentence is predicted as simple, which is a critical error in our scenario.  However, allowing false positives means the classifier unnecessarily flags simple sentences.  To mitigate this, we use the Area Under the ROC Curve (AUC) as the primary evaluation metric, as it is less sensitive to class imbalance.  For example, the degenerate classifier which only predicts the majority class will achieve the AUC score of 0.5, reflecting the absence of meaningful discrimination \cite{li24auc}.

\subsection{Models and Training Setup}

Our experiments use the multilingual BERT-base model \cite{devlin19}, as well as multilingual SBERT for sentence-level embedding representations \cite{reimers2019sentence}. Preliminary experiments with mBERT-large and XLM-RoBERTa \cite{conneau20emerging} revealed similar performance trends; therefore, we focus on the base architectures for clarity and consistency. 

Baseline mBERT models were fine-tuned independently on the English and French datasets using the Hugging Face Transformers library, trained for up to ten epochs with early stopping. The English corpus serves as a standard benchmark, while the larger French dataset (nearly two million sentences) enables analysis of how corpus size influences model robustness to noise. For evaluating cross-lingual transfer we applied the English and French models to Catalan, Spanish, Italian, and Russian: Catalan is a low-resource language for PLM pre-training, so we expect weaker results, while Russian is distant to English and French, thus these languages provide interesting cases for assessing the multilingual transfer gap \cite{hu2020xtreme}. Since the PLMs were pre-trained primarily on English, stronger English performance is expected. However, the French Vikidia dataset is much larger, so it allows us to test whether data scale can compensate for pre-training bias. 


\subsection{Noise Reduction Pipeline}

We define noise as either incorrect labels or low-quality text. The incorrect labels are Vikidia sentences that are actually complex or Wikipedia sentences that are simple, as distinct from correctly labeled sentences (either simple or complex), which we seek to keep.  Low-quality text concerns data-extraction artifacts, such as list-like or malformed segments, encoding issues or markup.

Therefore, we applied several noise reduction methods to identify unreliable samples in the training data with the aim of removing them to avoid annotation inconsistencies and to improve prediction accuract. After detecting noisy instances, only samples identified as clean were retained for fine-tuning, using the same hyperparameters as the baselines to ensure comparability. For evaluation, we manually cleaned the test sets (removing/repairing clear mislabels or broken segments) to ensure reliable measurement; results on the original noisy test are lower, as expected.

In addition to assessing each denoising method individually, we conducted an \textit{intersection analysis} to identify sentences consistently flagged as noisy across multiple approaches, revealing the most recurrent noisy examples and their effect on downstream performance.    
  
We compare five denoising methods:
\vspace{-1.3ex}
\paragraph{Gaussian Mixture Models (GMMs).}
GMMs cluster high-dimensional sentence representations into two distributions corresponding to clean and noisy samples \cite{bishop2006pattern}. We experimented with representations from BERT CLS embeddings (GMM-B) and from SBERT (GMM-SB). Initially, hyperparameters—such as the number of components, covariance type, and threshold—were tuned using Optuna over 100 trials \citep{akiba2019optuna}, targeting maximal separation between clean and noisy clusters. Based on these trials, we selected two fixed, cross-lingual configurations to standardise experiments and reduce computational costs, so that we set the number of GMM components to 9 for both languages and models to keep the clustering consistent and stable. For SBERT, we used \emph{full} covariance matrices because its embeddings generates more semantically meaningful embeddings using a siamese architecture and pooling strategies \citep{reimers2019sentence}. In contrast, BERT [CLS] embeddings are unevenly distributed in space and can lead to unstable clusters if full covariance is used. Therefore, the best model uses \emph{tied} covariance for BERT, which shares the same shape across all clusters and improves stability \citep{ethayarajh2019contextual}. We set the noise threshold by applying kernel density estimation to the GMM noise scores and selecting the dip between peaks \citep{silverman1981multimodality}. This dynamic thresholding strategy captures data-specific separation between noisy and clean points and adapts to the inherent variability of each corpus.
We apply this unified GMM configuration across English and French to ensure a consistent modeling approach, enhancing reproducibility and comparability across multilingual settings.

\paragraph{Small-Loss Trick (ST).}
ST assumes that examples with high training losses are more likely to contain noise \cite{arpit2017closer,han2018co}. During training, the model retains only a subset of samples with lower losses to detect data points likely to be mislabeled \cite{yu2019does}.  A key hyperparameter in this method is the \emph{loss threshold percentile}, which determines the fraction of lowest-loss examples retained at each selection step: at each epoch, losses are computed for all current examples, for example, a 75\% threshold thus means that at each selection point we retain the bottom 75\% of examples ranked by loss and exclude 25\% with the highest loss.  For the next epoch another 75\% of the lowest losses from the full dataset are selected for training.  After testing the range from 10\% to 90\% we have selected the 75\% threshold for the best balance, effectively excluding uncertain data points while retaining a sufficient number of reliable instances for model training. 
Finally, we define \emph{noisy} sentences as those repeatedly assigned to the excluded (high-loss) set across the five epochs (i.e., the intersection of high-loss selections over epochs).



\begin{table*}[t]
    \centering
    \small
    \setlength{\tabcolsep}{2pt}
    \renewcommand{\arraystretch}{1.1}
    \begin{subtable}[t]{\linewidth}
    \resizebox{\linewidth}{!}{%
    \begin{tabular}{l|ccccccc|cccccc}
\hline

 &  &  &  &  &  &  &  & \multicolumn{6}{c}{\textbf{Intersection}} \\
\textbf{Language} 
& \textbf{Baseline} 
& \textbf{GMM-B} 
& \textbf{GMM-SB} 
& \textbf{ST} 
& \textbf{CT} 
& \textbf{NTM} 
& \textbf{LS} 
& \textbf{CT/LS} 
& \textbf{LS/NTM} 
& \textbf{CT/NTM} 
& \textbf{CT/G-S} 
& \textbf{LS/G-S}
& \textbf{CT/NTM/G-S} \\
\hline
\hline
en & 0.5209 & \underline{0.9206} & \underline{0.9211} & 0.8296 & 0.7790 & 0.8343 & 0.8116 & \underline{0.9235} & \underline{0.9259} & 0.9114 & 0.9096 & \underline{0.9196} & \textbf{0.9261} \\
\hline
ca & 0.5099 & 0.7693 & 0.7504 & 0.7239 & 0.7494 & 0.7277 & 0.8003 & 0.7764 & 0.7558 & 0.7692 & 0.7560 &0.7605&0.7525  \\
es & 0.5180 & 0.7753 & 0.7760 & 0.7509 & 0.7724 & 0.7407 & 0.7955 & 0.7770 & 0.7750 & 0.7806 & 0.7528 & 0.7697&0.7735 \\
fr & 0.5207 & 0.7736 & 0.7733 & 0.7255 & 0.7358 & 0.7262 & 0.6612 & 0.7859 & 0.7735 & 0.7760 & 0.7637 &0.7678 &0.7702 \\
it & 0.5104 & 0.7718 & 0.7699 & 0.7502 & 0.7259 & 0.7212 & 0.7974 & 0.7714 & 0.7719 & 0.7738 & 0.7628 &0.7699 &0.7678\\
ru & 0.5447 & 0.7968 & 0.8015 & 0.7301 & 0.7899 & 0.7976 & 0.8475 & 0.7797 & 0.7982 & 0.7917 & 0.7756  &0.8146&0.8067  \\
\hline
\textbf{Average} 
& 0.5207
& \underline{0.7774}
& \underline{0.7742}
& 0.7361
& 0.7547
& 0.7427
& \textbf{0.7804}
& \underline{0.7781}
& \underline{0.7749}
& \underline{0.7783}
& 0.7622
& \underline{0.7765}
&  \underline{0.7741} \\

 Train (s) & 19,977 & 25,536 & 13,658& 22,706 & 36,723 & 24,239 & 30,348 & 74,291 & 62,627 & 75,457 &58,443 & 52,471& 89,115 \\
\hline
\# Noisy & -- & 3037 & 37070 & 24,525 & 157,979 & 72,128 & 72,678 & 476 & 14,612 & 31,328 &16,200  &7,429& 3,215\\
\# Percent & -- & 0.84\% & 10.20\% & 6.75\% & 43.47\% & 19.85\% & 20.00\% &0.13\%  & 4.02\%& 8.62\% & 4.46\% & 2.04\%&0.88\% \\
\hline
\%noisy Wiki
& --
& 98.58\%
& 93.88\%
& 100\%
& 88.99\%
& 94.26\%
& 94.33\%
& 99.58\%
& 94.55\%
& 89.34\%
& 88.57\%
& 94.45\% &88.52\%\\

\%noisy Viki
& --
& 1.42\%
& 6.12\%
& 0\% 
& 11.01\%
& 5.74\%
& 5.67\%
& 0.42\%
& 5.45\%
& 10.66\%
& 11.43\%
& 5.55\%&11.48\% \\
\hline
\end{tabular}
}
\subcaption{Trained on English}
\label{tab:noise-en}

\end{subtable}

\vspace{0.75em}
\begin{subtable}[t]{\linewidth}
    \resizebox{\linewidth}{!}{%
   \begin{tabular}{l|ccccccc|cccccc}
\hline
 &  &  &  &  &  &  &  & \multicolumn{6}{c}{Intersection} \\
Language & Baseline & GMM-B & GMM-SB & ST & CT & NTM & LS 
& \textbf{CT/LS} & \textbf{LS/NTM} & \textbf{CT/NTM} & \textbf{CT/G-S} & \textbf{LS/G-S}& \textbf{CT/NTM/G-S} \\
\hline
fr 
& 0.9198 & \underline{0.9213} & 0.9193 & 0.9182 & 0.5432 & 0.7694 
& \textbf{0.9402} & 0.9240 & \underline{0.9265} & 0.9216 & 0.9221 &\underline{0.9235} & \underline{0.9238} \\
\hline

en 
& 0.8523 & 0.8585 & 0.8623 & 0.8480 & 0.5045 & 0.7530 
& 0.8516 & 0.8296 & 0.8430 & 0.8344 & 0.8332 &0.8360& 0.8411 \\

ca 
& 0.7677 & 0.7660 & 0.7628 & 0.7801 & 0.5089 & 0.6836 
& 0.7697 & 0.7550 & 0.7704 & 0.7722 & 0.7607 &0.7585& 0.7700 \\

es 
& 0.8033 & 0.8090 & 0.8044 & 0.8047 & 0.5033 &0.7167 
& 0.7991 & 0.7909 & 0.7996 & 0.7899 & 0.7898 &0.7940& 0.7913 \\

it 
& 0.8076 & 0.8137 & 0.8123 & 0.8127 & 0.5086 & 0.6982 
& 0.8080 & 0.7940 & 0.8072 & 0.7928 & 0.7931 &0.7972& 0.7969 \\

ru 
& 0.7783 & 0.7807 & 0.7670 & 0.7587 & 0.4946 & 0.7443 
& 0.7478 & 0.7456 & 0.7675 & 0.7618 & 0.7439 &0.7533& 0.7632 \\
\hline

\textbf{Average} 
& \underline{0.8018}
& \textbf{0.8056}
& \underline{0.8018}
& \underline{0.8008}
& 0.5040
& 0.7192
& \underline{0.7952}
& 0.7830
& \underline{0.7975}
& 0.7902
& 0.7841
& 0.7878
& \underline{0.7925} \\

Train (s) &58,386  &69,490  & 110,424 & 93,637 & 80,342 & 174,600  & 69,899  & 243,540 & 320,370  & 330,813 &270,111 &259,931& 441,237  \\
\hline
\# Noisy & -- & 1,108,670  &  349,441 & 434,377 &747,823   & 279,472 & 566,356 &183,487  & 56,468 & 81.349 &40,732 &25,144 & 30,482\\
\# Percent & -- & 48.96\% & 15.43\% & 19.18\% & 33.01\% & 12.34\% & 25.02\% & 8.10\% & 2.49\% &3.59\% &2.09\% &1.11\% & 1.35\%\\
\hline
\%noisy Wiki
& --
& 88.87\%
& 86.19\%
& 89.14\%
& 69.08\%
& 85.78\%
& 86.04\%
& 65.55\%
& 83.83\%
& 66.17\%
& 70.41\%& 85.59\%
& 69.56\% \\

\%noisy Viki
& --
& 11.13\%
& 13.81\%
& 10.86\%
& 30.92\%
& 14.22\%
& 13.96\%
& 34.45\%
& 16.17\%
& 33.83\%
& 29.59\%& 14.41\%
& 30.44\%\\

    \hline
\end{tabular}
    }
\subcaption{Trained on French}
\label{tab:noise-fr}
\end{subtable}
    \caption{ Comparative performance of noise reduction methods (measured via ROC-AUC) with models trained on English (top) and French (bottom).
``Average'' is the macro-average ROC-AUC over transfer languages (excluding the training language).
``Baseline'' refers to standard fine-tuning with no noise filtering. \# Noisy is the count of sentences detected as noisy by the respective method. G-B stands for GMM-Bert, G-S stands for GMM-SBert.  The underlined values are not significantly different from the best ones.
}
    \label{tab:noise-both}
\end{table*}

\paragraph{Co-Teaching (CT).}
CT extends the Small-loss Trick method by training two prediction models in parallel, each selecting the lowest-loss samples from the other’s batch, under the assumption that these are more likely to be correctly labeled \cite{han2018co}. This cross-filtering mechanism ensures that each model learns from cleaner examples, reducing the influence of noisy data.  The key hyper-parameter in CT is the \textit{forget rate}, which determines the fraction of high-loss samples to remove from training at each training step.
Dynamic Loss Thresholding (DLT) enables the model to adapt gradually, preventing premature discarding of difficult samples and reducing excessive data loss in early training. This outperforms static thresholding by enhancing robustness against noisy labels \cite{yang2021dlt}.  In our implementation, we used a dynamic forget rate schedule, increasing linearly from 0.0 to 0.3 across epochs:
\begin{equation}
    r_t = r_{\min} + (r_{\max} - r_{\min}) \frac{t}{T},
\end{equation}
\noindent
where \( r_t \) is the forget rate at epoch \( t \), \( r_{min} = 0.0 \) is the initial forget rate, \( r_{max} = 0.3 \) is the maximum forget rate, and \( T \) is the total number of epochs. At each training step, the forget rate determines the fraction of high-loss samples to discard within each mini-batch. In the early epochs, all samples are used (\( r_t = 0 \)), while in later epochs, up to 30\% of the higher-loss samples are discarded (\( r_t \to 0.3 \)).

\paragraph{Noise Transition Matrix (NTM).}
NTMs explicitly model the probability of label corruption under class-dependent noise by estimating each matrix element \(T_{ij} = P(\tilde{y}= j \mid y = i)\), where \(y\) and \(\tilde{y}\) are the true and observed labels, respectively \citep{patrini2017making}.  Unlike CT and ST, which discard noisy samples based on loss, noise transition-based methods retain all data and adjust model predictions to compensate for systematic label noise. 
Using data identified as noisy by GMM-SB, we estimated the matrix \(T_{ij}\) from the empirical confusion matrix and computed its inverse \(T_{\text{inv}}\). During training, predicted probabilities \(\hat{P}\) are adjusted as:
\begin{equation}
\hat{P}_{\text{adjusted}} = \hat{P} \cdot T_{\text{inv}}
\end{equation}
followed by cross-entropy loss computation. This procedure retains all samples while correcting for systematic label noise.

\paragraph{Label Smoothing (LS).}
Label smoothing regularises the model by softening categorical targets. Instead of assigning a probability of 1 to the correct class and 0 to all others, LS redistributes a small fraction of this probability across all classes, helping the model handle mislabeled or ambiguous data \cite{szegedy2016rethinking,muller2019does}. This reduces overconfidence and has shown to improve generalisation on noisy or imbalanced datasets \cite{lukasik2020does,ren2024learning}. The smoothed label is computed as:
\begin{equation}
y_{\text{smooth}} = (1 - \epsilon) \times y + \frac{\epsilon}{k},
\end{equation}
\noindent
where \( \epsilon\)  is the smoothing factor and \( k \) is the number of classes. The noise-rejection threshold \( \tau \) was tested in the range \( 0.50 \leq \tau \leq 0.70 \), incremented by \( 0.05 \), while the smoothing factor was varied between \( 0.0 \leq \epsilon \leq 0.2 \), also incremented by \( 0.05 \). 

The results indicate that higher smoothing factors (\(\epsilon \geq 0.15\)) excessively redistributed probabilities, leading to degraded predictions due to increased uncertainty in class assignments. Conversely, the absence of smoothing (\(\epsilon = 0.0\)) resulted in overconfident predictions, increasing the risk of misclassification. A moderate smoothing factor of \( \epsilon = 0.1 \) provided the optimal balance, enhancing generalisation while maintaining well-calibrated predictions. Similarly, higher noise-rejection thresholds (\(\tau = 0.70\)) yielded superior performance by effectively filtering uncertain predictions.

\begin{figure*}[t]
    \centering
    \begin{subfigure}{0.49\linewidth}
        \centering
        \includegraphics[width=\linewidth]{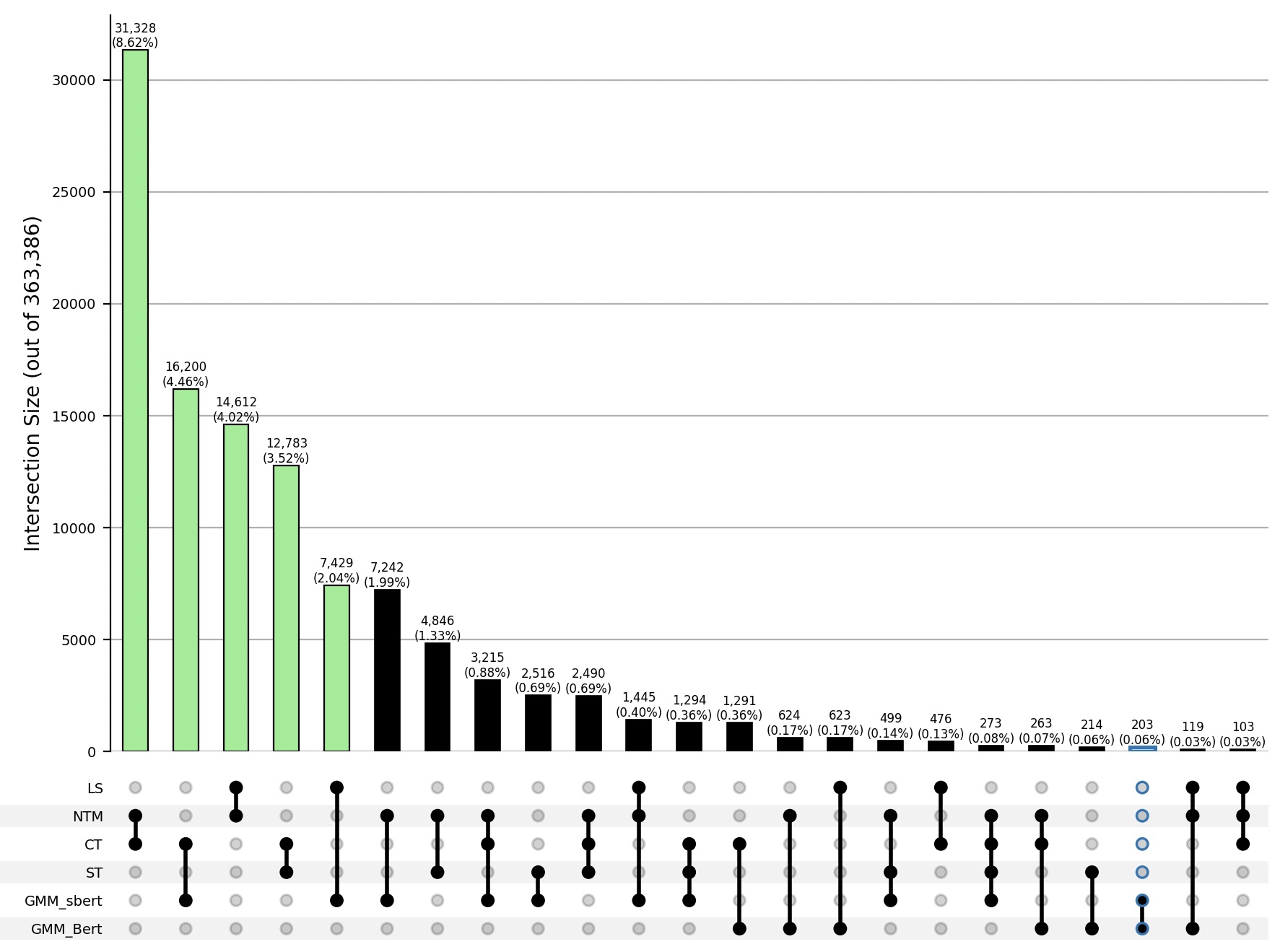}
        \caption{Noisy data intersections for English. }
    
        \label{fig:upset_en}
    \end{subfigure}
    \hfill
    \begin{subfigure}{0.49\linewidth}
        \centering
        \includegraphics[width=\linewidth]{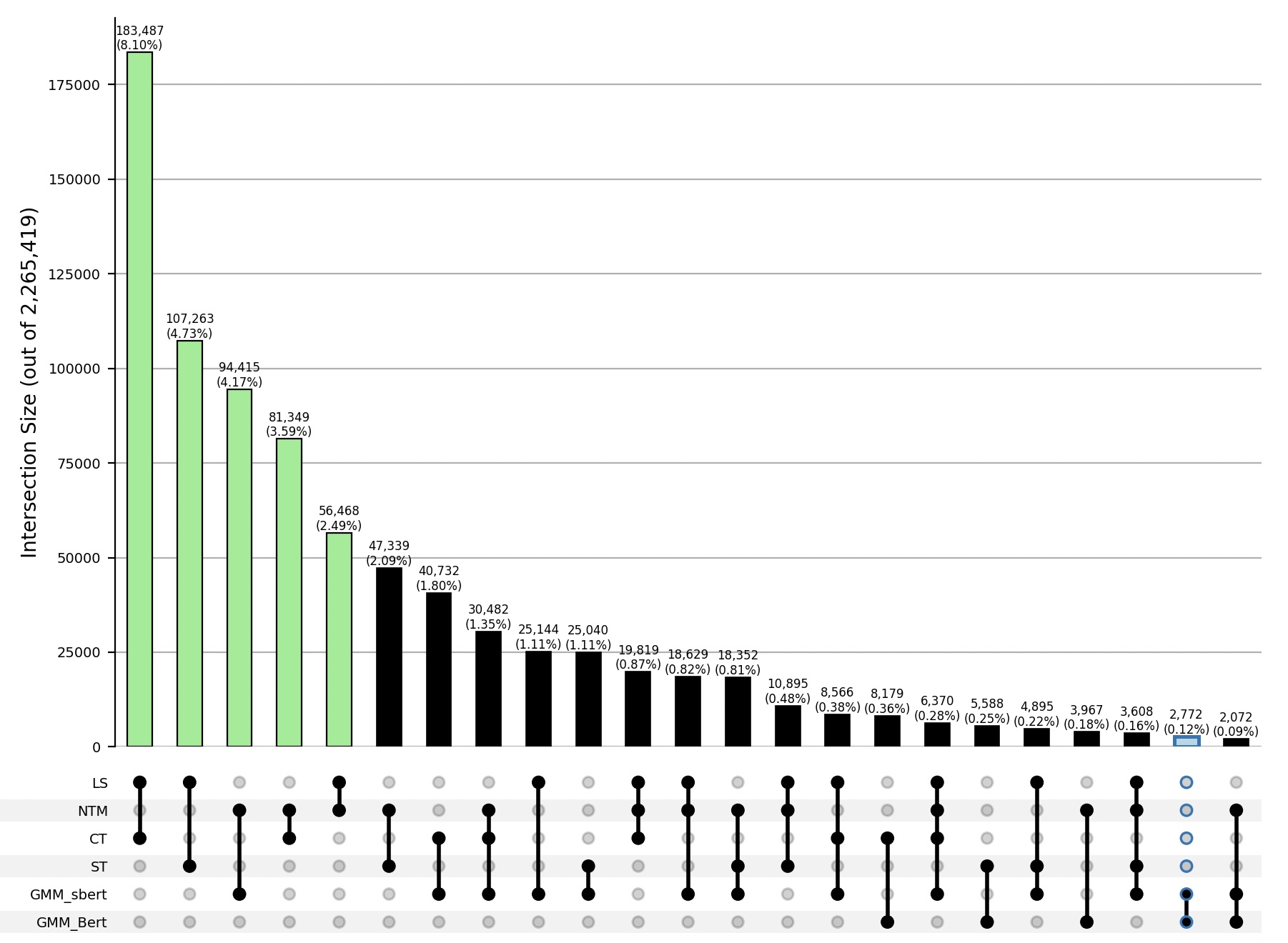}
        \caption{Noisy data intersections for French.}
    
        \label{fig:upset_fr}
    \end{subfigure}
    \caption{Comparison of noisy data intersections across denoising techniques for English and French. Each vertical bar represents the proportion of sentences identified as noisy by the combination of methods in the connected dots below. The small bar chart at the bottom left indicates the contribution of each method to the total pool of sentences identified as noisy.}
    \label{fig:upset_comparison}
\end{figure*}

\section{Noise Detection and Reduction}
\label{sec:noise_reduction}
\subsection{Impact de-noising on performance}
Our evaluation focuses on the impact of denoising techniques on sentence-level classification robustness, particularly in cross-lingual settings. Our goal is to assess how effectively different noise-reduction strategies enhance model stability and transfer performance across languages, using English and French as the primary training datasets. 
The experiments use the sentence-level datasets described in \autoref{tabData}.

When trained on English (Table~\ref{tab:noise-both}\subref{tab:noise-en}), the baseline model achieved a very poor AUC value, which has been substantially improved by all de-noising methods.  Both Gaussian Mixture Models (on the BERT embeddings and on the Sentence-BERT embeddings) achieved higher and more stable discrimination.  As expected, cross-lingual transfer decreases the performance.  However, de-noising can keep it to acceptable values with the best method of GMM-Bert and GMM-SBert and Label Smoothing.  

The number of sentences flagged as noisy varies widely across methods: from 3,037 for GMM-Bert (less than 1\% of the original dataset) to over 157,000 for CT (43\%), which leads to a substantially less accurate classifier trained on CT-filtered data.  This indicates that the de-noising methods have different sensitivities. 

Beyond the individual de-noising methods, we experimented with their intersection, i.e. by removing the data detected as noisy by two classifiers. The resulting intersection identifies smaller but more reliable subsets of noise.  The best-performing intersections---CT/LS and LS/NTM (0.93 for English, 0.78 for the cross-lingual transfer)--demonstrate that integrating even relatively weak denoising methods can yield fairly balanced outcomes.
However, the computational costs are more significant for computing the intersection; for example, computing both CT and GMM-SBert on the English dataset using a single L40S GPU adds 43,395 seconds to the time of fine-tuning the baseline model (19,977 sec).

In contrast to English, models trained on the much larger French dataset (Table~\ref{tab:noise-both}\,\subref{tab:noise-fr}) exhibited a different trend. The baseline already achieved the level of performance comparable with the best English model after de-noising. De-noising for French improves over the baseline model only marginally, namely from 0.9189 to 0.9402 for the single method (LS) or to 0.9265 for the intersection (LS/NTM), the same de-noising techniques which worked well for English. This indicates diminishing returns from denoising when training data is abundant. The close linguistic proximity between French and the Romance test languages (Catalan, Spanish, Italian) helps in achieving better cross-lingual transfer from French in comparison to Russian, which has seen no benefits from the bigger French dataset. 
   
From a computational efficiency perspective, both training and de-noising become more expensive for French.  While the classification models provide slightly better performance on French and in cross-lingual transfer, the training time increased, both for the baseline model (58,386 sec) and for the best de-noising (69,899 sec).  At the same time, the de-noising step helps with removing less relevant examples, so a combination of de-noising with training is less that the sum of de-noising and baseline training. 

\subsection{Intersection analysis}
We also investigated the agreement across the denoising techniques through their intersection (Figure~\ref{fig:upset_comparison}), which shows the size of intersections, i.e., sentences jointly flagged by the methods indicated by the connected dots. For the English-trained model \ref{fig:upset_en}, noise detection is dominated by combinations involving CT, particularly with NTM (8.62\%), GMM-SBERT (4.46\%), and ST (3.52\%), indicating that CT consistently overlaps with other methods in identifying noisy samples. The LS–NTM intersection (4.02\%) also contributes substantially, suggesting that loss-based and transition-based modelling approaches converge with probabilistic clustering methods on similar subsets of noisy data. These high-overlap regions represent the most reliable noise detections, as multiple independent techniques agree on the same examples.


The analysis shows a clear concentration of cross-method agreement in noise around CT- and NTM-based methods, with GMM-SBert appearing consistently across several high-agreement intersections and contributing to multiple overlapping regions, alongside smaller yet meaningful overlaps involving LS and ST.

For the French-trained model \ref{fig:upset_fr}, noise detection is likewise dominated by combinations involving CT, particularly with LS (8.10\%), the ST (4.73\%), and NTM (3.59\%), 
indicating that CT consistently overlaps with other methods in identifying difficult or mislabeled samples. The NTM–GMM-SB intersection (4.17\%) also contributes substantially, suggesting that probabilistic clustering and loss-based modelling converge on similar subsets of noisy data. These high-overlap regions represent the most reliable noise detections, as multiple independent techniques agree on the same examples. The analysis shows a clear concentration of noise consensus around LS and CT, with additional meaningful overlaps involving NTM and GMM-SB. 
This pattern suggests that CT and LS form a robust backbone for noise identification in the large-scale French dataset, while NTM and GMM-based methods provide complementary filtering signals. Overall, these intersections confirm that multi-method agreement serves as a strong indicator of genuine annotation noise, 
improving both reliability and interpretability of the denoising process.

\begin{table}
\small
\begin{tabularx}{\linewidth}{X r r}
\toprule
\textbf{Category} & \textbf{English} & \textbf{French} \\
\midrule
\textbf{Total sentences} & 230 & 237 \\
\textbf{Not noisy} & 92 & 78 \\
\textbf{Noisy} & 138 & 159 \\
\bottomrule
\end{tabularx}
\caption{Accuracy analysis at the intersection}
\label{tabAccuracy}
\end{table}

\section{Manual Error Analysis}
\label{sec:error-analysis}
To validate our denoising pipeline, we analyse errors in two stages: (i) selecting a subset of sentences flagged jointly by CT, NTM, and GMM-SBert to check whether there are reasons to treat them as noisy, and (ii) reviewing the noisy sentences to develop a noise taxonomy. 
Guided by Table~\ref{tab:noise-both}, we select the three-way intersection (CT$\cap$NTM$\cap$GMM-SB) as a high-confidence subset of noisy predictions. This intersection ranks among the strongest-performing intersections in both English- and French-trained settings. This subset captures agreement between three complementary signals: loss-based filtering (\textit{CT}), explicit modelling of label transitions (\textit{NTM}), and embedding-based clustering (\textit{GMM-SB}). As a result, it is less likely to include sentences that are simply difficult but clean, because an instance must be supported by all three mechanisms to be retained. Moreover, this intersection represents the strongest multi-method agreement observed consistently across both English- and French-trained settings. For the sentences in which there was a reason to treat them as noise, we have also performed analysis of the possible categories.  This has led to detecting three main categories: \emph{Structural} noise (artifacts which lack clausal structure), \emph{Content} noise (distributionally atypical for the classifier), and \emph{Label} noise.  The subcategories of \emph{Structural} noise have been defined as:

\begin{description}
    \item[SF:] Structural fragments (truncation / broken segmentation), e.g.,
    \emph{``The dwarf planet Ceres is by far the largest asteroid, with a diameter of .''};
    \emph{``, 85 HTCs have been observed, compared with 664 identified JFCs.''}

    \item[MS:] Markup or symbolic artifacts (wiki/template spillover, formulas/symbols), e.g.,
    \emph{``value:rgb (1,0.7,0.7) \; Legend:Xbox\_360\_...''};\\
    \emph{``+ 4 CO $\rightarrow$ 3 Fe + 4 \ldots C + HO $\rightarrow$ CO + H.''}

    \item[EN:] Enumerations and list spillovers (category/tag lists), e.g.,
    \emph{``Apple Inc.|1976 establishments in California|American brands|\ldots|Steve Jobs|Technology companies \ldots''};
    \emph{``Game Boy Color games|Mario platform games|\ldots|Virtual Console games for Wii U|\ldots''}

\item[LA:] Language switch or encoding (non-English scripts or Unicode characters embedded inline), e.g.,
\emph{``(Hong Kong) Co., Ltd.\ (\cjk{松下信興機電 (香港)有限公司}) and Panasonic SH\ldots''};
\emph{``The word physics comes from the Greek word ἡ φύσις (`nature').''}

    \item[CI:] Citation or reference fragments (dangling citations, references, or file/link spillover), e.g.,
    \emph{``\textuparrow{} `Sky \& Telescope: March 2008', Southern Hemisphere Highlights \ldots Allen, R.\ H.\ (1899) \ldots''};
    \emph{``Internet map 1024.jpg \textbar{} Partial map of the Internet \ldots Structure of the Universe.jpg \textbar{} Galactic\ldots''}
\end{description}

The subcategories of \emph{Content} noise are:
\begin{description}
\item[NE:] Density of named entities, e.g.,
\emph{``Cover athlete: Kak\'a (World), Wayne Rooney (United Kingdom), Mesut \"{O}zil \& Ren\'e Adler (Germany), Tim Cahill (Australia) \ldots''}

\item[NU:] Numerical or measurement density, e.g.,
\emph{``In currency, there are pennies (\$0.01), nickels (\$0.05), dimes (\$0.1), quarters (\$0.25), half dollars (\$0.5), and dollar coins (\$1).''};

\item[TE:] Technical terminology density, e.g.,
\emph{``\ldots Theoretical and Experimental Nuclear Physics, Nonlinear Optics, Thin film Magnetism, Neutron Scattering, Neutron Activation Analysis \ldots''}

\item[AC:] Acronym or abbreviation density, e.g.,
\emph{``Consumers are given the option to have any URL ending in .weebly.com, .com, .net, .org, .co, .info, or .us.''}

\end{description}

The \emph{Label} noise category covers two cases: sentences from Wikipedia are simple to be included in Vikidia (coded as MC) and sentences from Vikidia should have been considered complex (coded as MS).

\autoref{tabAccuracy} shows the contribution of errors to the sentences detected as noisy.  Even though, the sample was selected as agreement between noise detection methods, there was no reason to suspect noise in 38.7\% of them for English and 33.1\% for French.  However, Table~\ref{tab:noisy_analysis} shows that the rate of mislabelling (0.200 for English, 0.262 for French) in the case of what has been detected as noise was far higher than the error rate of the respective classifiers (0.043 for English and 0.083 for French).  In practical terms, this means that a substantial portion of the flagged sentences in both languages are incorrect labels, which is the reason for improvement in performance once they are removed. 

The most common category in both languages is the presence of structural artifacts from data processing, such as fragments, markup residue, list spillovers, or truncated segments. A second frequent source of noise is label noise, where sentences appear well-formed but assigned with the wrong complexity label (46 in English; 62 in French). These cases likely arise from annotation inconsistencies when document-level judgements are projected onto individual sentences. Content noise is the least common category, where sentences are dominated by standalone lists of names, numbers, or domain terminology (30 in English; 41 in French). These cases are often well-formed, but they are out-of-distribution for sentence-level difficulty prediction: when a line is mostly a list of entities or technical terms, it lacks normal sentence structure and its dense token pattern can be mistaken for high complexity.

While structural artifacts constitute the largest share of flagged noise, the presence of wrongly labeled sentences is particularly critical. These cases reflect annotation inconsistencies caused by projecting document-level complexity labels onto individual sentences.
These findings provide strong motivation for applying denoising methods,
as filtering or down-weighting such instances reduces the risk of learning incorrect patterns and improves the reliability of sentence-level difficulty classification.

\begin{table}[t]
\centering
\small
\renewcommand{\arraystretch}{1.1}
\begin{tabularx}{\linewidth}{X r r}
\toprule
\textbf{Category} & \textbf{English} & \textbf{French} \\
\midrule

\textbf{Structural}: Total & \textit{185} & \textit{186} \\
\midrule
SF:Structural fragment & 59 & 83 \\
MS:Markup or symbolic artifact & 46 & 36 \\
EN:Enumeration or list artifact & 44 & 57 \\
LA:Language switch or encoding  & 17 & 2 \\
CI:Citation or reference fragment & 19 & 8 \\

\midrule
\textbf{Content}: Total & \textit{30} & \textit{41} \\
\midrule
NE:Named entity density & 14 & 9 \\
NU:Numerical or measurement density & 9 & 2 \\
TE:Technical terminology density & 3 & 25 \\
AC:Acronym or abbreviation density & 4 & 5 \\

\midrule
\textbf{Label:} Total & \textit{46} & \textit{62} \\
\midrule
MS: Mislabeled as Simple & 22 & 24 \\
MC: Mislabeled as Complex & 24 & 38 \\

\midrule

\textbf{Total sentences} & 230 & 237 \\
\bottomrule
\end{tabularx}
\caption{Manual analysis of noise for English and French. Counts reflect annotation assignments (multiple labels per sentence allowed). Totals may exceed the number of noisy segments.}
\label{tab:noisy_analysis}
\end{table}
\begin{figure*}[t]
\centering
\begin{subfigure}[t]{0.48\linewidth}
    \centering
    \includegraphics[width=\linewidth]{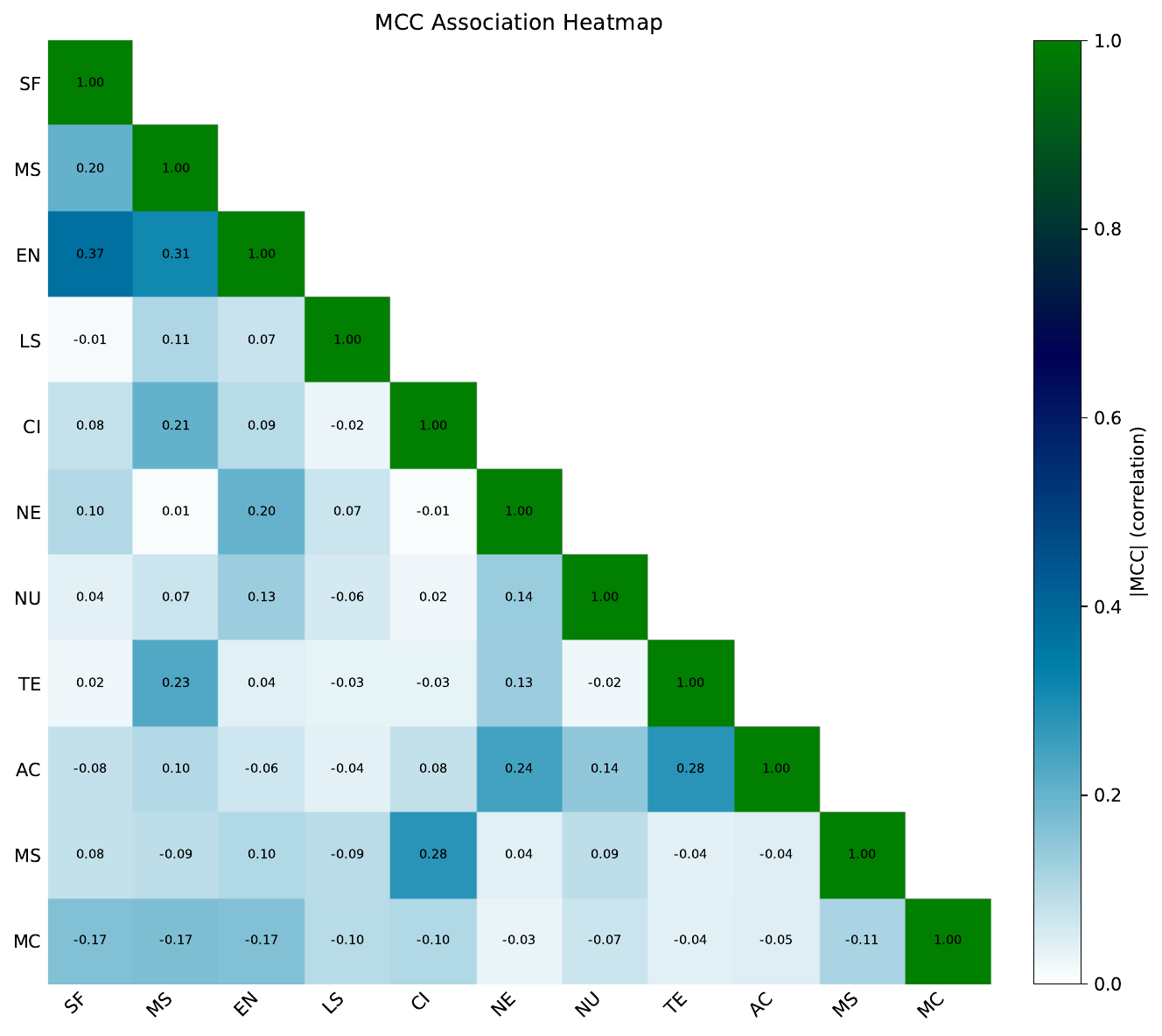}
    \caption{English}
    \label{fig:en_mcc_heatmap}
\end{subfigure}
\hfill
\begin{subfigure}[t]{0.48\linewidth}
    \centering
    \includegraphics[width=\linewidth]{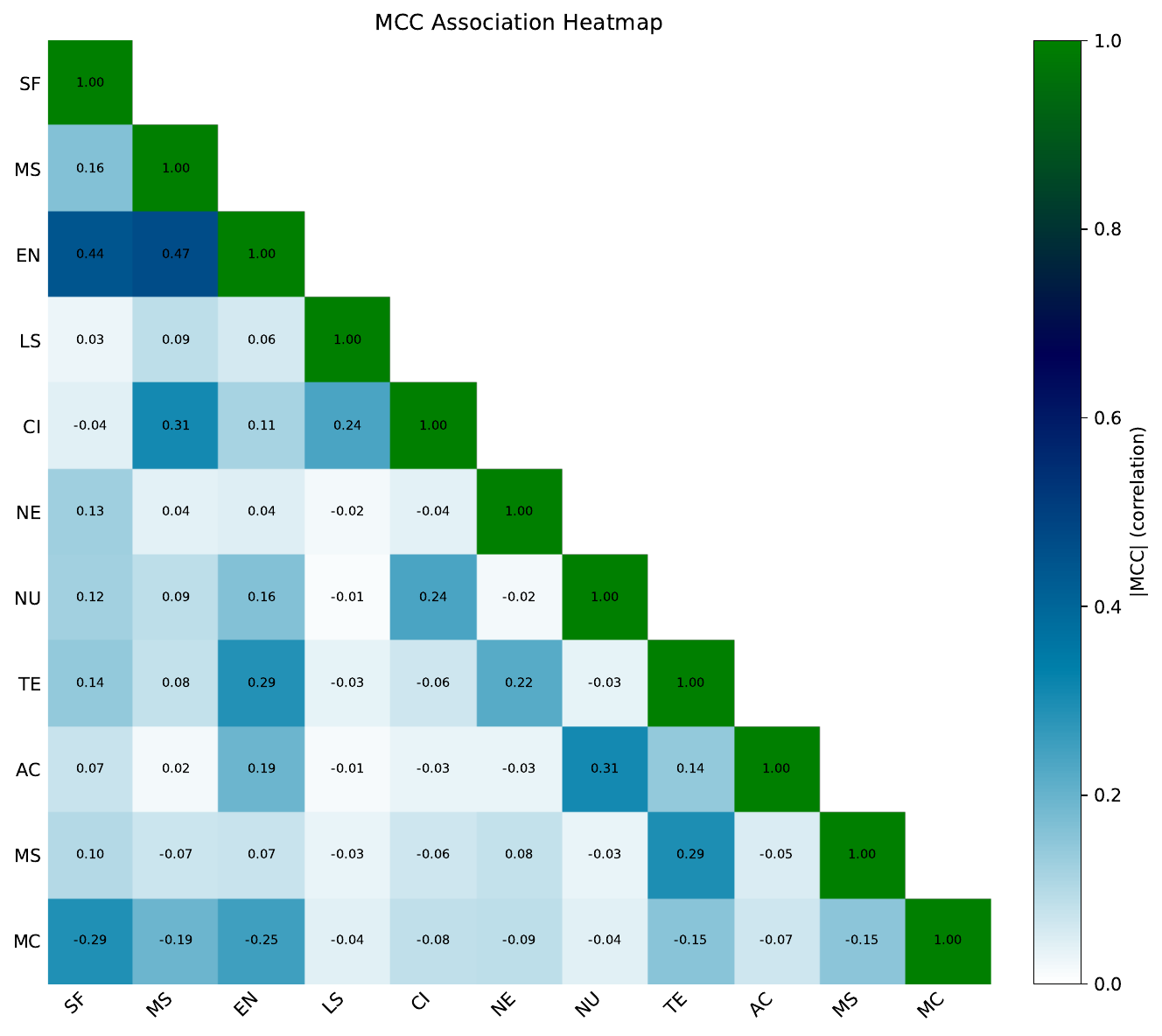}
    \caption{French}
    \label{fig:fr_mcc_heatmap}
\end{subfigure}

\caption{Pairwise associations between noise categories measured using the Matthews Correlation Coefficient (MCC) for English (left) and French (right).}
\label{fig:mcc_heatmaps}
\end{figure*}

To further investigate how the annotated noise categories interact, Figure~\ref{fig:mcc_heatmaps} presents the pairwise Matthews Correlation Coefficient (MCC) \cite{matthews1975comparison} between subcategories for English and French. While Table~\ref{tab:noisy_analysis} reports category frequencies, the MCC heatmaps reveal patterns of co-occurrence, indicating which noise phenomena tend to appear together within the same sentence.

Across both languages, the strongest positive associations are concentrated among the structural noise. In particular, \textit{SF}, \textit{MS}, and \textit{EN} tend to co-occur, with the highest correlations appearing in French (peaking at MCC $=0.47$ for \textit{MS--EN}, and $0.44$ for \textit{SF--EN}). This suggests that “noise” is often not a single isolated issue: markup spill-over and formatting artifacts frequently trigger segmentation failures and fragment-like outputs, creating a compact structural core of errors. In contrast, content-density noise appears more independently. These categories (e.g., \textit{NE}, \textit{AC}, \textit{TE}) only co-occur in a few specific pairs—such as \textit{TE--AC} and \textit{NE--AC} in English—rather than showing strong links across the whole taxonomy. Similarly, \textit{LA} is near zero with most other tags.

A key observation is that \textit{MC} (label noise) has mostly weak, and often negative, correlations with the structural cluster (e.g., as low as $-0.29$ with \textbf{SF} in French). This suggests that many mislabelled instances are actually well-formed sentences: they are problematic mainly because their assigned complexity label is incorrect, not because the text is corrupted in the usual sense.
Overall, these results show that supervision errors in both languages come from different sources: structural issues, density-related effects, and annotation inconsistencies. This supports the need for denoising methods that can handle different types of errors, rather than assuming that all noise follows the same pattern.


The qualitative patterns help explain the complementary behaviour of our methods:
\begin{enumerate}
    \item GMM-based filters mostly detect \emph{distributional outliers} (domain NE density, symbolic tokens), yielding high-precision removal of atypical segments.
    \item CT/ST preferentially downweight \emph{structural fragments and corrupted strings} that produce unstable or persistently high losses across epochs.
    \item LS does not \emph{detect} noise per se, but improves calibration under mixed uncertainty (e.g., residual encoding noise or context loss), reducing overconfidence without aggressive data deletion.
\end{enumerate}
These observations align with our quantitative results: intersection between GMMs and Co-Teaching captures the most consistent artifacts, while Label Smoothing achieves competitive reliability at low computational cost.

Content noise categories are often linked to the lack of discourse context. This supports our earlier finding that longer segments improve robustness by reducing under-contextualised cases.

\section{Related Studies}

Noise in training data poses a significant challenge in NLP, especially in non-topical classification tasks such as genre prediction \cite{ronnqvist22explaininggenres,roussinov23emnlp}, demographic property detection \cite{kang19dataset}, and text difficulty classification \cite{north22complexity}. These tasks rely on language style rather than explicit topical keywords, making them sensitive to noise and annotation errors.

Noise reduction techniques like majority agreement between the classifiers have been effective. Studies by \citet{dibari14} and \citet{khallaf21difficulty} show that leveraging consensus between the predictions of different models can significantly reduce noise, resulting in more reliable classifiers. Additionally, \citet{zhu22robust} provide a baseline by evaluating BERT models' robustness to label noise, without a clear outcome on which denoising methods are more useful. Our study goes further, by selecting a non-topical classification task and real-life settings by shifting prediction from document- to sentence-level predictions.

Bayesian learning has also been applied to handle noise, as discussed by \citet{papamarkou24bayesian} and \citet{miok20semisup}, focusing on managing uncertainty and noise in large-scale AI tasks. This approach is particularly relevant for semi-supervised text annotation, where it enhances noise reduction efficacy. Given the amount of unlabeled data in our domain, we will apply the experiments to the Bayesian framework.

Calibration of model predictions is crucial for handling noise, particularly using softmax outputs. Proper calibration ensures lower probabilities correspond to a higher likelihood of errors, aiding in producing well-calibrated classifiers. Methods for uncertainty estimates in BERT-like models can improve robustness to noise at the inference stage \cite{vazhentsev23uncertainty,khallaf26uncertainty}, which we need to investigate further.



Cross-lingual transfer learning, where models like BERT are trained on one language and applied to another, is particularly challenging in noisy environments due to linguistic differences and resource variability \cite{conneau20emerging,zhao21fewshot}.

\section{Conclusions}
\label{sec:org8c39538}
This paper examined the impact of various noise reduction techniques on cross-lingual sentence difficulty classification, providing insights into their effectiveness across languages and datasets. Our findings show that noise reduction can improve model performance, although its effectiveness depends on dataset characteristics and cross-lingual transfer conditions.

In our smaller dataset, Gaussian Mixture Models (GMMs) proved effective in mitigating noise. By contrast, in our larger dataset, the inherent regularisation properties of pretrained language models provide a strong baseline, with more computationally intensive denoising methods yielding only marginal additional gains.  However, having a cleaner dataset reduces the training efforts and overall helps with future experiments.

These findings have practical implications for improving the robustness of cross-lingual applications in domains such as language education, text simplification, and language learning tools. By tailoring noise reduction strategies to dataset size and structure, developers can enhance the reliability and interpretability of sentence complexity models across languages.

The released dataset is the largest multilingual dataset for language difficulty prediction on the sentence level.

\section{Acknowledgments}
This document is part of a project that has received funding from the European Union’s Horizon Europe research and innovation program under Grant Agreement No. 101132431 (iDEM Project).  The University of Leeds was funded by UK Research and Innovation (UKRI) under the UK government’s Horizon Europe funding guarantee (Grant Agreement No. 10103529).  The views and opinions expressed in this document are solely those of the author(s) and do not necessarily reflect the views of the European Union. Neither the European Union nor the granting authority can be held responsible for them.

\section{Limitations}
This study aims at a specific non-topical classification task, for which there have been no prior experiments on denoising. The specific setup of moving from document- to sentence-level annotation relies on available Vikidia-Wikipedia pairs and on the availability of annotators, which limited the number of languages for testing. Our definition of difficulty relies on the Vikidia/Wikipedia proxy and, despite test-set cleaning, may encode corpus-specific biases beyond difficulty.
Future work will explore the applicability of these denoising techniques to other multilingual datasets and classification tasks, particularly in low-resource settings and for multiclass classification scenarios, as well as for other non-topical classification tasks.  Additionally, investigating alternative sources of sentence-level annotations or adapting methods for diverse text genres (e.g., social media, news, or educational content) could further assess the robustness of our approach.  

\section{Ethical Impact}
The potential societal benefits of our findings are substantial, particularly in improving the quality of communication by detecting complex sentences across languages. This study will also contribute to production of cleaner de-noised datasets.

In conducting the study we have been careful with the environmental impact of NLP research. Large Language Models are more computationally expensive, while they have been shown to be not better than BERT-like PLMs in text classification tasks. For each of the methods we estimated the computational costs of running the models (on NVIDIA L40S GPUs), with a total training time of \(\approx 19\) hours for English and \(\approx 201\) hours for French.
We are not aware of potential risks in deploying the study discussed in the paper.

\section{Bibliographical References}\label{sec:reference}

\bibliographystyle{lrec2026-natbib}
\bibliography{bibexport}

\appendix

\end{document}